\documentclass[conference]{IEEEtran}
\IEEEoverridecommandlockouts
\usepackage{cite}
\usepackage{amsmath,amssymb,amsfonts}
\usepackage{algorithmic}
\usepackage{graphicx}
\usepackage{textcomp}
\usepackage{xcolor}

\usepackage{multirow}
\usepackage{bbding}
\def\BibTeX{{\rm B\kern-.05em{\sc i\kern-.025em b}\kern-.08em
    T\kern-.1667em\lower.7ex\hbox{E}\kern-.125emX}}
\begin{document}

\title{Exploiting Inductive Bias in Transformer for Point Cloud  Classification and Segmentation}

\author{\IEEEauthorblockN{1\textsuperscript{st} Zihao Li, 2\textsuperscript{nd} Pan Gao}
\IEEEauthorblockA{
\textit{College of Computer Science and Technology}\\
\textit{Nanjing University of Aeronautics and Astronautics, Nanjing}\\
pride\_19@163.com, Pan.Gao@nuaa.edu.cn}
\and
\IEEEauthorblockN{3\textsuperscript{rd} Hui Yuan}
\IEEEauthorblockA{
\textit{School of Control Science and Engineering} \\
\textit{Shandong University, Jinan}\\
huiyuan@sdu.edu.cn}
\and
\IEEEauthorblockN{4\textsuperscript{th} Ran Wei$^{\ast}$ \thanks{*Corresponding author}}
\IEEEauthorblockA{\textit{Science and Technology on Electro-optic Control Laboratory, Luoyang} \\
115946873@qq.com}
\and
\IEEEauthorblockN{5\textsuperscript{th} Manoranjan Paul}
\IEEEauthorblockA{
\textit{Charles Sturt University, Bathurst}\\
mpaul@csu.edu.au}
\and
}

\maketitle

\begin{abstract}
Discovering inter-point connection for efficient high-dimensional feature extraction from point coordinate is a key challenge in processing point cloud. Most existing methods focus on designing efficient local feature extractors while ignoring global connection, or vice versa. In this paper, we design a new Inductive Bias-aided Transformer (IBT) method to learn 3D inter-point relations, which considers both local and global attentions. Specifically, considering local spatial coherence, local feature learning is performed through Relative Position Encoding and Attentive Feature Pooling. We incorporate the learned locality into the Transformer module. The local feature affects value component in Transformer to modulate the relationship between channels of each point, which can enhance self-attention mechanism with locality based channel interaction. We demonstrate its superiority experimentally on  classification and segmentation tasks. The code is available at: https://github.com/jiamang/IBT
\end{abstract}


\section{Introduction}
Point cloud is a digitalized data of the surface on a real three-dimensional object, containing geometry coordinates, and sometimes, also having attribute information. The original point cloud data can be obtained by equipment such as lidar or depth camera. Because of its irregularity and disorder, new challenges have been brought to point cloud processing.

Thanks to Pointnet \cite{qi2017pointnet} for opening a precedent for directly absorbing raw point cloud data \cite{wu20153d,uy2019revisiting,yi2016scalable}, and also thanks to many public point cloud datasets that are available for training and evaluation, methods based on deep learning continuously emerge. Some approaches \cite{qi2017pointnet,qi2017pointnet++,wu2019pointconv,thomas2019kpconv,li2018pointcnn} based on MLP or convolution operations can directly operate on unstructured raw points for end-to-end training. However, these methods lack the modeling of the interconnection between points. Considering the inherent topology of unstructured data, many studies \cite{wang2019dynamic,te2018rgcnn,li2019deepgcns} have applied graph convolutional neural networks to point cloud learning and achieved good results.
 The graph structure can establish correlation between points and extract local information between nodes. But how different neighbor points contribute to the center point is not investigated. In recent years, inspired by the success of the attention mechanism in the NLP field, many works \cite{guo2021pct,zhao2021point,yan2020pointasnl,zhang2022patchformer,mazur2021cloud,zhong2021point} have also applied it to point cloud to further improve network performance. It mainly calculates and obtains local or global attention coefficients, and then linearly combines them with the corresponding feature vectors to obtain output features, which can distinguish the different levels of contributions  of each point. But those attentions are paid to improving the ability of local feature extractors while ignoring global attention and the interaction between local and global attentions.

To address these issues, we propose an efficient Inductive Bias-aided Transformer (IBT). Inductive Bias is indeed like the local consistency of CNN filters, which is mentioned in many papers \cite{dosovitskiy2020image,battaglia2018relational,goyal2022inductive}. Broadly speaking, inductive bias encourages learning algorithms to prioritize solutions with certain properties. IBT mainly consists of three crucial modules.
\begin{figure*}[t]
    \centering
   \includegraphics[width=17cm,height=9cm]{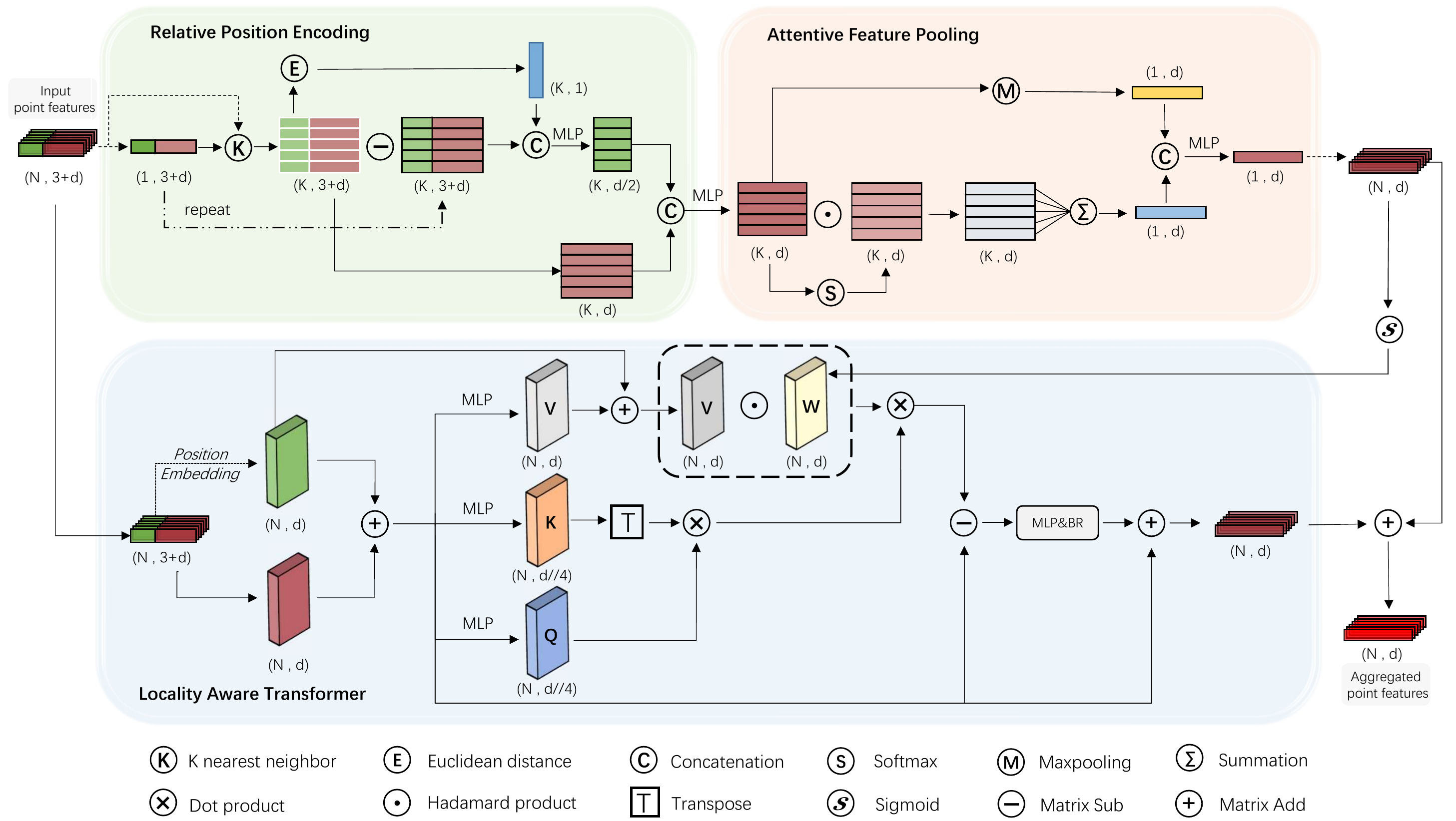}
   \caption{The proposed Inductive Bias-aided Transformer. The top-left shows the relative position encoding module for extracting position information, and the top-right illustrates the procedure of fusion of two types of feature pooling. The bottom panel shows how the improved transformer utilizes local features to enhance self-attention mechanism.}
\label{module}
\vspace{-4mm}
\end{figure*}

The Relative Position Encoding module is mainly to explicitly encode the spatial layout of 3D points, which plays a crucial role in shape analysis. Existing methods 
\cite{qi2017pointnet++,zhao2019pointweb,wang2019dynamic} usually concatenate location information and features into features for learning, but are not ideal in capturing meaningful ensemble patches. Therefore, the advantage of our design module is to first explicitly encode the point position, and then combine the output with the feature of the point to further enhance it, so that each point can observe its local geometry, thus finally enriching the entire network. This lays a good foundation for subsequent deep feature learning. Attentive Feature Pooling in the follow-up design is to update the features of the center point, focusing on deep relational learning. we designe two branches to fully integrate deep-level features, including maximum pooling to obtain the most prominent features of neighboring points and attention pooling to automatically select the most important features. In order to establish the connection between local and global features, we pass the learned local features through the sigmoid function to obtain the weight coefficient of each channel for each point, and integrate them into the Locality Aware Transformer for global feature learning.
We stack three layers of IBT to obtain semantic information from coarse-to-fine for better classification and segmentation. We have done extensive experiments on the ModelNet40 \cite{wu20153d}, ScanObjectNN \cite{uy2019revisiting} and ShapeNetPart \cite{yi2016scalable} datasets to confirm its effectiveness. Our main contributions are summarized as follows:

\begin{itemize}
    \item We design an effective Inductive Bias-aided Transformer layer to deeply integrate local attention with global attention. The inductive bias refers specifically to the locality used to model local visual structure. 

    \item We design Relative Position Encoding and Attentive Feature Pooling to obtain the most critical and representative local features, and improve the traditional transformer with locality-based channel interaction.
 
    \item By incorporating the above modules into the design of the network architecture, our method surpasses previous best performing work in point cloud classification, and yields the best IoU performance for most of object categories, demonstrating very competitive segmentation performance. 
\end{itemize}

\section{Related work}
Direct processing on raw points will not damage original geometry information. PointNet \cite{qi2017pointnet} was the first of its kind to advance this type of research and achieved good results. It mainly uses MLP for each point independently and aggregates the global features with a symmetric function. Subsequent development work can be roughly divided into three categories: MLP or convolution based, graph convolution-based and Transformer-based methods.

\emph{MLP or Convolution based Methods.} It extracts features through MLP or convolution. Since PointNet \cite{qi2017pointnet} ignores the interconnection between points, the subsequently proposed PointNet++ \cite{qi2017pointnet++} uses a multi-scale structure to effectively capture local features. In addition, there are some studies on how to perform convolution on points. For example, PointConv \cite{wu2019pointconv} introduces point density into the convolution kernel. This algorithm also proposes a strategy for effectively calculating weight functions to increase network depth and performance.

\emph{Graph-based Methods.} It updates features by constructing local or global graphs. Dynamic graph-based DGCNN \cite{wang2019dynamic} is good at capturing geometric structural features from constructed local graphs. In addition, DeepGCN \cite{li2019deepgcns} connects all points as a global graph structure, introduces residual connection and convolution into graph convolutional network (GCN), which finally builds a GCN as deep as CNN.

\emph{Transformer-based Methods.} It pays more attention to important parts by computing different coefficients between points. For example, PCT \cite{guo2021pct} obtains deeper semantic information by stacking several global offset attention layers. PT \cite{zhao2021point} uses attention to obtain semantic information of different resolutions via the process of progressive downsampling. In addition, PointANSL \cite{yan2020pointasnl} employs the attention mechanism to handle noisy point cloud data via adaptive sampling.

\begin{figure*}[h]
    \centering
   \includegraphics[width=16cm,height=7cm]{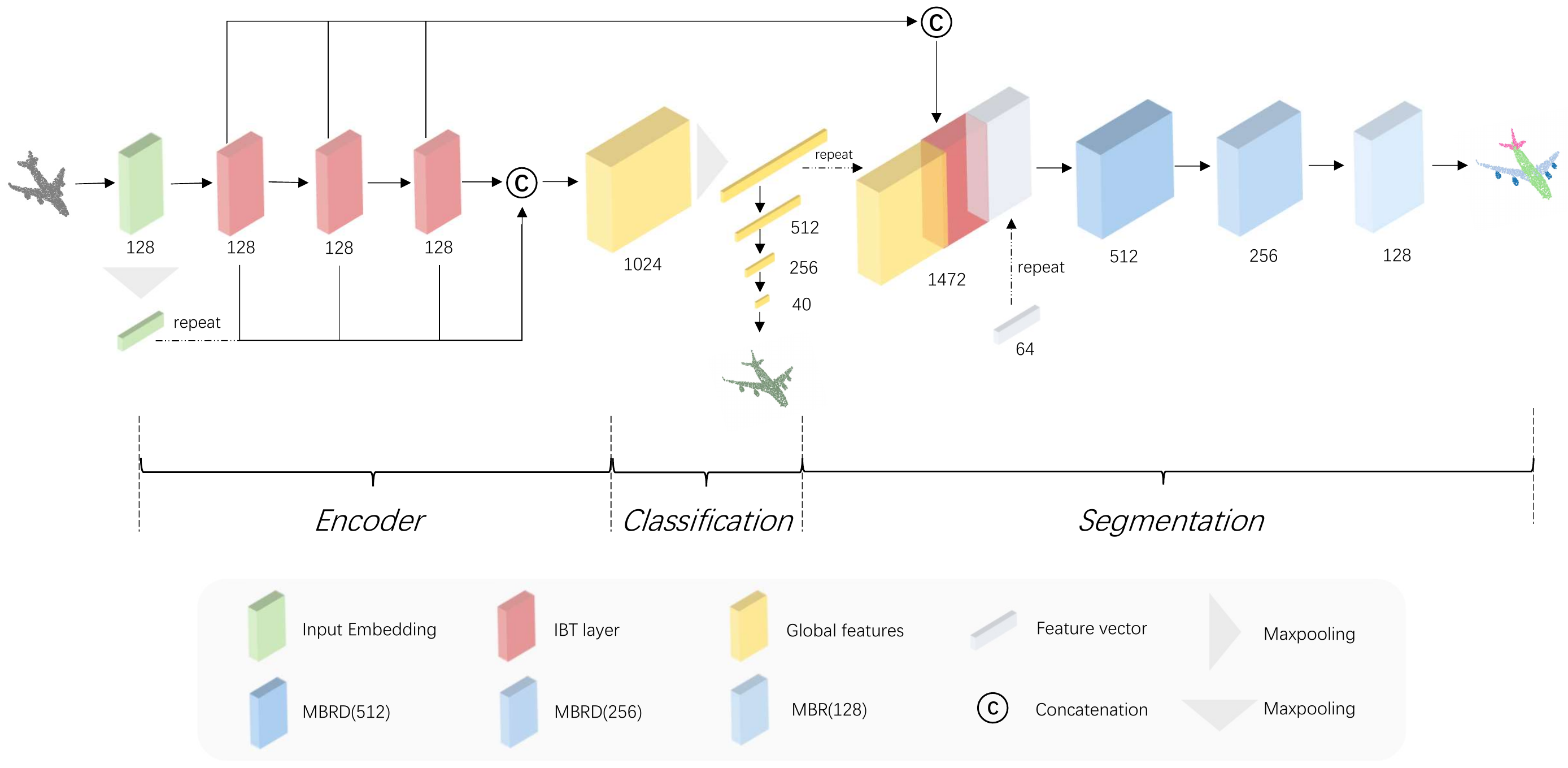}
   \vspace{-5pt}
   \caption{Network architecture for point cloud classification and segmentation tasks.}
\label{architecture}
   \vspace{-8pt}
\end{figure*}

\vspace{-0.3cm}

\section{Method}
 We define the original input point cloud as $X=\left\{x_i|i = 1,2,...,N\right\}\in\mathbb{R}^{N\times3}$ . Here $x_i$ represents the three-dimensional coordinates $(x,y,z)$ of the {\emph i}-th point. Similarly, we define the feature of each point as $F=\left\{f_i|i = 1,2,...N\right\}\in\mathbb{R}^{N\times D}$ . We use KNN to find k-nearest neighbors for each center point to construct a local graph structure, including self-loops, defined as $\mathcal{G}=(\mathcal{V},\mathcal{E})$, where $\mathcal{V}=\left\{1,2,...K\right\}$ and $\mathcal{E}\subseteq \mathcal{V}\times\mathcal{V}$ are the {\em vertices} and {\em edges}, respectively. Suppose that $x_i$ is the center point of the graph structure, and then $N(i)=\left\{j:(i,j)\in\mathcal{E}\right\}$ is the neighboring point set. Next, we will introduce our Inductive Bias-aided Transformer and the network architecture for point cloud classification and segmentation.
\subsection{Local Feature Extraction}
We construct a local graph by finding k-nearest neighbors for each center point in space, and then update features by interacting information with neighboring points. We define the feature of the edge as $e_{ij} = h_\Theta(f_i,f_i^j)$ , where $h_\Theta:\mathbb{R}^D \times \mathbb{R}^D \rightarrow \mathbb{R}^{M}$ is a nonlinear function with a set of learnable parameters $\Theta$. Finally, the features of all edges are aggregated to obtain local features. As shown in the upper part of Figure \ref{module}, it is mainly implemented by Relative Position Encoding and Attentive Feature Pooling.

{\bf Relative Position Encoding.} For each centroid point, we encode its relative spatial position with k-nearest neighbors and their Euclidean distance to better capture local geometric patterns. Additionally, we retrieve and encode feature differences between pairs of points in the local graph. Finally, they are concatenated and subjected to MLP operation to obtain rich relative position information. The process is written as follows:
\begin{equation}
p_{xi} = MLP ((x_i-x_i^j)\oplus\parallel x_i-x_i^j\parallel\oplus (f_i-f_i^j))
\end{equation}
where $j\in N(i)$. $x_i$ and $x_i^j$ represent the original three-dimensional coordinates, $f_i$ and $f_i^j$ represent the corresponding features, $\oplus$ is the concatenate operation, and $\parallel \cdot \parallel$ calculates the Euclidean distance between the neighbours and center point.
Next, we concatenate the location information and point features into  Feature Pooling module after changing the dimension through MLP. The expression is as follows:
\begin{equation}
h_{xi} = MLP (p_{xi}\oplus f_i^j),~j\in N(i)
\end{equation}

\begin{figure*}[t]
    \centering
   \includegraphics[width=16cm,height=7cm]{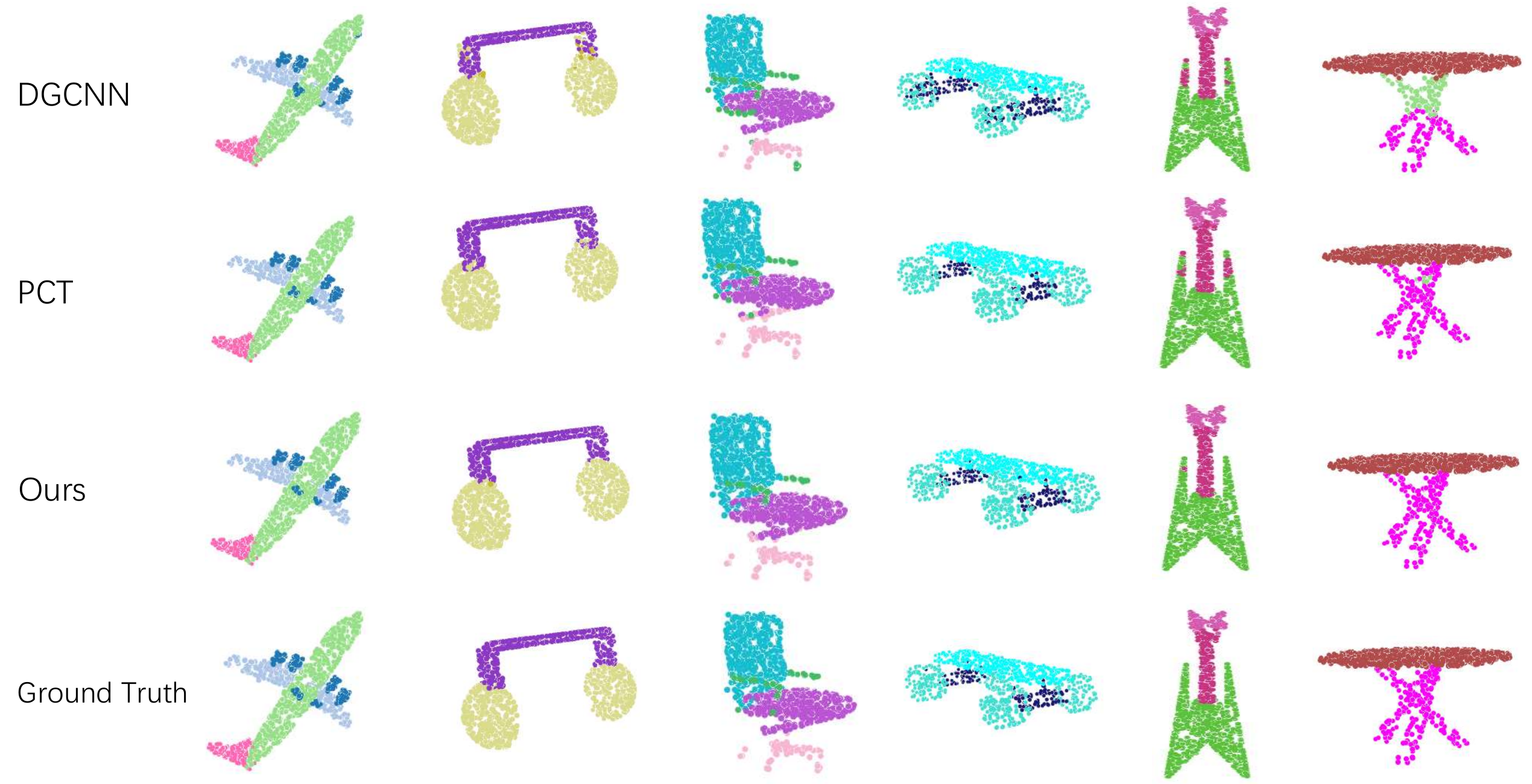}
   \vspace{-8pt}
   \caption{Visual comparison with other methods for part segmentation.}
\label{visual_p}
\vspace{-4mm}
\end{figure*}

{\bf Attentive Feature Pooling.} This module is used to aggregate features of neighboring points. We use a combination of max pooling and attention mechanism to integrate update features. Max pooling can preserve the most salient features of adjacent points, while the attention mechanism can automatically learn the importance of local features. 

Given a set of features of local graph points $F_i=\left\{f_i^1,f_i^2,...,f_i^j\right\}$, we design a shared function $g( )$ which consists of a shared MLP and a softmax to learn an unique attention score for each feature. It is defined as follows:
\begin{equation}
s_i^j = g (f_i^j,W_s)
\end{equation}
where $W_s$ is the learnable weight coefficient.

The learned attention score can automatically allocate the importances of  features for weighted summation. On the other hand, we also use the maximum pooling method to aggregate the features of neighbor points. For each center point, our Relative Position Encoding and Attentive Feature Pooling module learn to aggregate the geometric patterns and features of its K nearest points, and finally generates an informative feature vector $\hat{f}_i$ denoted as follows:
\begin{equation}
\hat{f}_i = MLP(\sum_{j=1}(f_i^j \cdot s_i^j) \oplus Max(f_i^j)), ~j\in N(i)
\end{equation}
\subsection{Locality Aware Transformer} Locality Aware Transformer integrates location embedding and local feature attention to improve traditional point cloud transformers. Positional embeddings allow the operator to adapt to local structures in the data, so each point is transformed to the same dimensionality as the features using a shared MLP. 
\begin{equation}
\delta = MLP(x_i)
\end{equation}
Then it is added with the features of the corresponding point as the starting feature $F_{in}$ as follows:
\begin{equation}
F_{in} = f_i + \delta
\end{equation}
In addition, we send the local features obtained in the previous section to a Sigmoid function for normalization to obtain a weight coefficient $W$ between 0 and 1 for each channel of each point as follows:
\begin{equation}
W=Sigmoid(\hat{F})
\end{equation}
Next, we use the attention mechanism to compute the semantic similarity between different items in a data sequence. Let $Q$, $K$, and $V$ be the \textit{query}, \textit{keyword}, and \textit{value} matrices generated by linear transformation on the input feature $F_{in} =\mathbb{R}^{N \times D} $  respectively, as follows:
\begin{equation}
Q,K=F_{in} \cdot (W_q,W_k)
\end{equation}
\begin{equation}
V= (F_{in} \cdot W_v + \delta) \cdot W
\end{equation}
where $W_q$, $W_k$, $W_v$ are shared learnable transformation parameters. We use the original self-attention mechanism to calculate the weight coefficients and multiply it with $V$. The process is expressed as follows:
\begin{equation}
F_{sa} = SoftMax(\frac{Q \times K^T}{\sqrt{d}}) \times V
\end{equation}
where $d$ is the dimension of query and key vector. In the experiment, it is set to D/4 to reduce the amount of calculation. Further, we employ the offset attention mechanism according to PCT \cite{guo2021pct}, which has strong theoretical basis of Laplacian operator. It can further improve network performance by computing the offset between self-attention features and input features with element-wise subtraction. The complete offset attention is expressed as follows:
\begin{equation}
F_{out} =MBR(F_{in} - F_{sa}) +F_{in}
\end{equation}
where MBR represents a combination of MLP, Batch Normalization and ReLU activation function.

\begin{table*}[t]
\caption{Part segmentation  on ShapeNetPart dataset. Metric is mIoU(\%), and NUM is the number of shapes  in the category.}
\begin{center}
\vspace{-4mm}
\resizebox{1\textwidth}{19mm}
{
\begin{tabular}{c|c|c|c|c|c|c|c|c|c|c|c|c|c|c|c|c|c}
\hline
Methods& mIou & air. & bag & cap  & car & cha. & ear.  & gui. & kni. & lam. & lap. & mot. & mug & pis. & roc. & ska.  & tab.\\
\hline
NUM &  & 2690 & 76 & 55 & 898 & 3758 & 69 & 787 & 392 & 1547 & 451 & 202 & 184 & 283 & 66 & 152 & 5271 \\
\hline
Pointnet\cite{qi2017pointnet} & 83.7 & 83.4 &78.7 & 82.5 & 74.9 & 89.6 & 73.0 & 91.5 & 85.9 & 80.8 & 95.3 &65.2 & 93.0 & 81.2 & 57.9 & 72.8 & 80.6 \\
Pointnet++\cite{qi2017pointnet++} & 85.1 & 82.4 &79.0 & 87.7 & 77.3 & 90.8 & 71.8 & 91.0 & 85.9 & 83.7 & 95.3 &71.6 & 94.1 & 81.3 & 58.7 & 76.4 & 82.6 \\
RGCNN\cite{te2018rgcnn} & 84.3 & 80.2 &82.8 & \textbf{92.6} & 75.3 & 89.2 & 73.7 & 91.3 & 88.4 & 83.3 & 96.0 &63.9 & 95.7 & 60.9 & 44.6 & 72.9 & 80.4\\
DGCNN\cite{wang2019dynamic} & 85.2 & 84.0 &83.4 & 86.7 & 77.8 & 90.6 & 74.7 & 91.2 & 87.5 & 82.8 & 95.7 &66.3 & 94.9 & 81.1 & 63.5 & 74.5 & 82.6\\
PCNN\cite{atzmon2018point} & 85.1 & 82.4 &80.1 & 85.5 & 79.5 & 90.8 & 73.2 & 91.3 & 86.0 & 85.0 & 96.7 &73.2 & 94.8 & 83.3 & 51.0 & 75.0 & 81.8\\
3D-GCN\cite{lin2020convolution} & 85.1 & 83.1 & 84.0 & 86.6 & 77.5 & 90.3 & 74.1 & 90.9 & 86.4 & 83.8 & 95.3 &65.2 & 93.0 & 81.2 & 59.6 & 75.7 & 82.8\\
MLMST\cite{zhong2021point}& 86.0 & 83.6 &\textbf{84.7} & 86.3 & 79.8 & 91.1  & 71.2 & 90.2   & \textbf{88.6} & 84.9 & 95.9  & 72.8  & 94.8 & 83.4   &56.2 & 76.7 & 82.6  \\ 
PointASNL\cite{yan2020pointasnl} & 86.1 & 84.1 &\textbf{84.7} & 87.9 & 79.7 & \textbf{92.2} & 73.7 & 91.0 & 87.2 & 84.2 & 95.8 &\textbf{74.4} & 95.2 & 81.0 & 63.0 & 76.3 & 83.2\\
PCT\cite{guo2021pct} & 86.4 & 85.0 &82.4 & 89.0 & \textbf{81.2} & 91.9 & 71.5 & 91.3 & 88.1 & \textbf{86.3} & 95.8 &64.6 & \textbf{95.8} & 83.6 & 62.2 & \textbf{77.6} & 83.7\\
\hline
Ours & 86.2 & \textbf{85.2} &81.4 & 86.1 & 80.1 & 91.5 & \textbf{76.6} & \textbf{91.9} & 87.6 & 84.6 &\textbf{97.1} & 72.9 & 95.4 & \textbf{84.3} & \textbf{63.7} & 76.5 &\textbf{83.9}\\
\hline
\end{tabular}}
\end{center}
\vspace{-10pt}

\label{segmentation result}
\vspace{-10pt}
\end{table*}

\subsection{Network Architecture} \vspace{-0.2cm} We design two network architectures for point cloud classification and segmentation tasks using the proposed Inductive Bias-aided Transformer (IBT) as shown in the figure \ref{architecture}. First, we send the point cloud coordinates to the input embedding layer using the shared MLP to obtain a higher-dimensional feature $F$, while saving the original coordinates. Then the coordinate $X$ and the feature $F$ of the corresponding point are sent to the three IBT layers to obtain detailed low-dimensional geometric features of different semantic levels. On the other hand, we max-pool the features obtained by the embedding layer to obtain a low-dimensional global feature vector, which can provide a coarse feature. In the experiments, the input embedding layer and the output features of the IBT layer are both 128-dimensional.

{\bf Classification network.} As shown in Figure \ref{architecture}, we concatenate three low-dimensional geometric features and global feature vectors, and upscale them to 1024 dimension through shared MLP. Finally, the scores of different prediction categories are output via the maximum pooling and linear layers.

{\bf Segmentation network.} Segmentation requires more detailed features, so on the basis of the 1024-dimensional global features obtained in the classification task, we also introduce the feature vector of the category and concatenate them with the detailed geometric features again, which are then gone through MLP, Batch Normalization, ReLU activation function and Drop layer (MBRD) to predict the category of each point.

\vspace{-0.2cm}
\section{Experiments}
\vspace{-0.2cm}
 We evaluate the effectiveness of the proposed IBT  using the ModelNet40 \cite{wu20153d} and ScanObjectNN\cite{uy2019revisiting} datasets for the point cloud classification and the ShapeNetPart \cite{yi2016scalable} dataset for the part segmentation. In addition, we also compare the performance with other deep learning and transformer-based methods to demonstrate its superiority.

{\bf Implementation details.} We implement IBT and point cloud learning tasks using Pytorch on two RTX 2080Ti GPUs. For classification and segmentation, we randomly sample the original point cloud into 1024 points and send them to the network for training and testing. Both use SGD optimizers with momentum and learning rate of 0.9 and 0.1, respectively. For the classification task, the k-nearest neighbor value is set to 40, and for the segmentation task, it is set to 80 in order to obtain more detailed features.

\vspace{-0.3cm}
\subsection{Classification} The ModelNet40 \cite{wu20153d} dataset contains 12311 CAD models in 40 categories. They are split into 9843 models for training and 2468 models for testing. For evaluation metrics, we use the average accuracy per class (mAcc) and the overall accuracy (OA) across all classes. It can be seen from the comparison in table \ref{classification result} with other mainstream methods that the method we proposed can obtain the highest average category and overall accuracy under the premise of only inputting 3D coordinates, which demonstrates its excellent performance.

\vspace{-3mm}
\begin{table}[h]
\caption{Classification results on ModelNet40.}
\begin{center}
\vspace{-2mm}
\scalebox{0.85}{\begin{tabular}{l c c c c}
\hline
Methods &Input & point & mAcc & OA\\
\hline
Pointnet++\cite{qi2017pointnet++}&xyz,normal & 5k & - & 91.9 \\
PointCNN\cite{li2018pointcnn}&xyz & 1k & 88.1 & 92.2\\
DGCNN\cite{wang2019dynamic}&xyz & 1k & 90.2 & 92.2\\
FPConv\cite{lin2020fpconv}&xyz,normal & 1k  & - & 92.5\\
PointConv\cite{wu2019pointconv}&xyz,normal & 1k & - &92.5\\
MLMST\cite{zhong2021point}&xyz & 1k &- &92.9\\
3DCTN\cite{xie2018attentional}&xyz,normal & 1k & 91.2 &93.3\\
PointASNL\cite{yan2020pointasnl}&xyz,normal & 1k & - &93.2\\
CloudTransformers\cite{mazur2021cloud}&xyz & 1k & 90.8 &93.1\\
PointStack\cite{wijaya2022advanced}&xyz & 1k & 89.6 & 93.3\\
PCT\cite{guo2021pct}&xyz & 1k & - &93.2\\
\hline
Ours &xyz & 1k& \textbf{91.0} & \textbf{93.6}\\
\hline
\end{tabular}}
\end{center}
\vspace{-13pt}

\label{classification result}
\end{table}

ScanObjectNN \cite{uy2019revisiting} is a realistic 3D point cloud classification dataset that is extremely challenging due to characteristics such as backgrounds, missing parts, and deformations. It consists of occluded objects extracted from real-world indoor scans and contains 2,902 3D objects from 15 categories. Some method results are taken from the official website test of the dataset. In addition, the number of input points and feature dimensions of the comparative methods are unknown, so they will not be listed here. The input of our experiment on this data set is only 3D coordinate  of 1024 points. We also compare the mean class and overall accuracy as shown in Table \ref{classification result2}. Due to the challenges of this dataset, the comparison with other methods highlights the efficiency and robustness of our method.
\vspace{-3mm}
\begin{table}[h]
\caption{Classification results on ScanObjectNN.}
\begin{center}
\vspace{-2mm}
\scalebox{0.85}{\begin{tabular}{l c c}
\hline
Methods& mAcc & OA\\
\hline
Pointnet++\cite{qi2017pointnet++}& 75.4 & 77.9 \\
DGCNN\cite{wang2019dynamic}& 73.6 & 78.1 \\
PointCNN\cite{li2018pointcnn}& 75.1 & 78.5 \\
BGA-DGCNN\cite{uy2019revisiting}& 75.7 & 79.7 \\
BGA-PN++\cite{uy2019revisiting}& 77.5 & 80.2 \\
DRNet\cite{qiu2021dense}& 78.0 & 80.3 \\
GBNet\cite{qiu2021geometric}& 77.8 & 80.5 \\
SimpleView\cite{goyal2021revisiting}&-&80.5\\
PRANet\cite{2021PRA}&79.1&82.1\\
\hline
Ours & \textbf{80.0} & \textbf{82.8}\\
\hline
\end{tabular}}
\end{center}
\vspace{-2mm}

\label{classification result2}
\end{table}

\subsection{Segmentation} The ShapeNetPart \cite{yi2016scalable} dataset consists of 16,880 models across 16 shape categories annotated for 3D object part segmentation. Among them, 14006 3D models are used for training and 2874 are used for testing. The number of parts in each category ranges from 2 to 6, for a total of 50 different parts. For each object, we convert the correct information of which category it originally belongs to into one-hot encoding, and after MLP, it is a 64-dimensional feature vector shown in Figure \ref{architecture}. For evaluation metrics, we report mIoU per class and mIoU across all shapes in all categories. The comparison with other mainstream methods is shown in  Table \ref{segmentation result}, and some visualization results are shown in Figure \ref{visual_p}. Our method works better on more categories, and it can be seen from the visual comparison that our method is closer to the ground truth.

\vspace{-0.2cm}

\subsection{Ablation experiment} In this subsection, we further explore each module of the proposed method and ablate different choices within the module. All ablation experiments are based on classification experiments with 1024 points on the ModelNet40 \cite{wu20153d} dataset. First, we ablate the main modules including Relative Position Encoding, Attentive Feature Pooling and Locality Aware Transformer modules as shown in the table \ref{ab_module}, where \Checkmark means to keep this module, blank means to remove this module, and replace it with MLP if necessary. It can be seen that the feature extraction only through the local feature extractor or only the global transformer is not as good as the proposed local and global interaction method. Forming weight coefficients by local features to enhance the global Transformer with channel attention can extract geometric detail features more comprehensively.

\vspace{-3mm}
\begin{table}[h]
\caption{Ablation experiments for different modules.}
\centering
\vspace{-2mm}
\scalebox{0.8}{
\begin{tabular}{c| ccc|cc} 
\hline
Model & ~Position & ~Pooling & Transformer & mAcc & OA  \\ 
\hline
A                         & \Checkmark         &\Checkmark        &             & 90.2  & 93.4   \\
B                         &           & \Checkmark      & \Checkmark           & 90.0  & 93.2   \\
C                         &           &          & \Checkmark           & 90.6  & 93.4   \\
D                         & \Checkmark         & \Checkmark        & \Checkmark          & 91.0  & 93.6   \\
\hline
\end{tabular}}

\label{ab_module}
\vspace{-2mm}
\end{table}

On the other hand, we explore the effect of removing some operations in each module and setting different k values as shown in Table \ref{ab_in}. It can be seen that if the k value is set too small, the interaction with the surrounding points is not enough. When the k value is too large, redundant information will be generated, which is not conducive to subsequent tasks. In addition, in the feature pooling module, it is proved that it is the most efficient way to combine the most significant features produced by the maximum pooling and the weighted features produced by the attention pooling. In the Locality Aware Transformer module, we also did ablation experiments on the weight coefficient $W$ used to link the local and global attentions, and whether the inclusion of location embedding has an impact. It can be seen that the architecture we have chosen  performs best.

\vspace{-3mm}
\begin{table}[h]
\caption{Ablation experiments for different options within the module.}
\centering
\vspace{-2mm}
\scalebox{0.8}{
\begin{tabular}{c|c|c |c} 
\hline
Module                                       &                       & mAcc& OA  \\ 
\hline
\multirow{4}{*}{Relative Position Encoding} & k=10                  & 90.0
 & 93.2  \\ 
\cline{2-4}
                                            & k=20                  & 90.2 & 93.3   \\ 
\cline{2-4}
                                            & k=40                  & 91.0  & 93.6  \\ 
\cline{2-4}
                                            & k=60                  & 90.6 & 93.0   \\ 
\hline
\multirow{2}{*}{Feature Pooling}            & w/o maxpooling        & 90.6  & 93.4  \\ 
\cline{2-4}
                                            & w/o attention pooling & 90.6 & 93.3   \\ 
\hline
\multirow{2}{*}{Locality Aware Transformer}                  & w/o weight $W$          & 90.5  & 93.3  \\
\cline{2-4}
&w/o Position Embedding &90.3&93.3 \\
\hline
\end{tabular}}

\label{ab_in}
\end{table}

\vspace{-0.5cm}

\section{Conclusion} This paper proposes a new transformer for 3D point cloud understanding, i.e., Inductive Bias-aided Transformer (IBT), and conducts extensive experiments to verify its effectiveness. The main contribution of our method is to extract local features and incorporates them into  an improved global Transformer in the form of channel clues. In this way, we can finally obtain a more detailed geometric feature represented by the fusion of local and global features. Our proposed model has a good performance on the public datasets both quantitatively and qualitatively. In addition, IBT is easy to integrate into many existing networks in a plug-and-play manner.

\section*{Acknowledgment}
\vspace{-2mm}
This work is supported in part by  the Natural Science Foundation of China under Grant 62272227 and the Aeronautical Science Foundation of China under Grant 201951052001.

\bibliographystyle{IEEEbib}
\bibliography{icme2023template}
\end{document}